%% file: elsarticle-template-1-num.tex
\documentclass[preprint,12pt]{elsarticle}

\usepackage{graphicx}
\usepackage{amssymb}
\usepackage{comment}

\usepackage{lineno}
\usepackage{amsmath}
\usepackage{booktabs}
\usepackage{multirow}
\usepackage{multicol}

\journal{XXX}

\begin{document}

\begin{frontmatter}


\title{MSP: An FPGA-Specific Mixed-Scheme, Multi-Precision Deep Neural NetworkQuantization Framework}



\author[a1]{Sung-En Chang}
\author[a1]{Yanyu Li}
\author[a1]{Mengshu Sun}
\author[a2]{Weiwen Jiang}
\author[a3]{Runbin Shi}
\author[a4]{Xuehai Qian}
\author[a1]{Xue Lin}
\author[a1]{YanzhiWang}

\address[a1]{Northeastern University}
\address[a2]{University of Notre Dame}
\address[a3]{University of Hong Kong}
\address[a4]{University of Southern California}

\begin{abstract}
With the tremendous success of deep learning, there exists imminent need to deploy deep learning models onto edge devices.
To tackle the limited computing and storage resources in edge devices, model compression techniques have been widely used to trim deep neural network (DNN) models for on-device inference execution.
This paper targets the commonly used FPGA (field programmable gate array) devices as the hardware platforms for DNN edge computing. 
We focus on the DNN quantization as the main model compression technique, since DNN quantization has been of great importance for the implementations of DNN models on the hardware platforms. 
The novelty of this work comes in twofold: (i) We propose a mixed-scheme DNN quantization method that incorporates both the linear and non-linear number systems for quantization, with the aim to boost the utilization of the heterogeneous computing resources, i.e., LUTs (look up tables) and DSPs (digital signal processors) on an FPGA. Note that all the existing (single-scheme) quantization methods can only utilize one type of resources (either LUTs or DSPs for the MAC (multiply-accumulate) operations in deep learning computations. (ii) We use a quantization method that supports multiple precisions along the intra-layer dimension, while the existing quantization methods apply multi-precision quantization along the inter-layer dimension. The intra-layer multi-precision method can uniform the hardware configurations for different layers to reduce computation overhead and at the same time preserve the model accuracy as the inter-layer approach.

Our proposed mixed-scheme, multi-precision (MSP) DNN quantization framework achieve $70.47\%$ Top1 accuracy in ResNet-18 on the ImageNet dataset. 
We also validate the proposed MSP framework on two FPGA devices i.e., Xilinx XC7Z020 and XC7Z045.
We achieve $3.53\times$ speedup in end-to-end inference time on the ImageNet, comparing with the fixed-point quantization method.

\end{abstract}


\end{frontmatter}


\input{Section/1_introduction}

\input{Section/2_background}
\input{Section/3_SP2}
\input{Section/4_MPQ}
\input{Section/5_results}
\input{Section/6_conclusion}






\bibliographystyle{elsarticle-num-names}
\bibliography{sample.bib}







\end{document}

%% file: Section/1_introduction.tex
\section{Introduction}

Deep neural networks (DNNs) have been employed in various of tasks with outstanding performance, such as convolutional neural networks (CNNs) for computer vision~\cite{lecun2015deep}, recurrent neural networks (RNNs) for natural language processing (NLP) ~\cite{bishop2006pattern,goodfellow2016deep}, etc. However, due to the large model size and extremely intensive computation, it is still challenging to deploy these DNN models on edge devices.

To support broad applications of deep learning, such as autonomous vehicles, wireless access points, robotic vision and control, smart health devices, etc., there are two aspects to enable DNNs under these resource constrained circumstances. First is to utilize specialized hardware platform for the inference of DNNs. Extensive research efforts have been dedicated to various kind of edge-computing platforms such as ASICs (Application Specific Integrated Circuits) ~\cite{mao2018lergan,hegde2018morph,han2016eie},  FPGAs ~\cite{sharma2016high,li2018network,zhang2015optimizing,shi2020ftdl}, and embedded CPUs/GPUs ~\cite{leng2019energy,niu2020patdnn,han2016mcdnn}. 

The other is DNN model compression which explores the potential of algorithm and hardware co-design and finds better trade-offs between task performance (accuracy, etc.) and hardware efficiency (latency, power consumption, etc.). There are in general two techniques for model compression: DNN pruning ~\cite{lym2019prunetrain,shi2020csb,han2015learning,liu2018rethinking} and quantization ~\cite{courbariaux2015binaryconnect}. These techniques can make models work with significantly smaller model sizes and fewer operations. Therefore, these compression techniques become a must-do step for deployment on edge devices.
Here we only focus on the quantization approach, which becomes imperative to hardware acceleration especially on the FPGA and ASIC platforms. By representing weights with fewer bits, weight quantization can directly reduce model size and accelerate inference speed.

In this paper, we propose the novel \emph{Mixed-Scheme, Multi-Precision (MSP) Quantization Framework} specifically for the FPGA devices, which achieves unprecedented hardware efficiency without sacrificing accuracy at low bit width (e.g., 4-bit). Our contributions are as follows: 
\begin{itemize}
    \item We develop a hardware-friendly sum-of-power-of-two (SPoT) quantization scheme, which can mitigate the accuracy degradation of the vanilla power-of-two (PoT) scheme while can still take advantage of bit shifting operation to accelerate computation.
    \item We propose a novel Mixed Scheme Quantization (MS) to fully utilize the heterogeneous resources of FPGAs. To be more specific, based on available DSP and LUT modules in different FPGAs, we can finalize the corresponding ratio of SPoT and fixed-point quantization scheme, so that both module can be fully exploited.
    \item Rather than using a different precision in the first and last layers, or employing inter-layer multi-precision for extreme compression rate, we propose an \emph{Intra-Layer Flexibility}, which can be applied to all layers in a DNN model, achieving lossless accuracy performance without damaging hardware efficiency.
\end{itemize}

%% file: Section/2_background.tex
\section{Related Work and Motivation}\label{sec:background}

In this section, we summarize quantization schemes and compare them in terms of their hardware deployment implications and accuracy degradation. Further, we also briefly discuss recent works on neural network quantization. 
Finally, we describe the motivation of our FPGA-specific MSP quantization.

\subsection{Quantization Schemes}

Based on whether the distances between the quantization levels are equal or not, there are linear and non-linear quantization schemes. In terms of bit width, we can also classify neural network quantization into catagories of single-precision or multi-precision. 

\subsubsection{Linear Number System}

Linear quantization schemes contain binary, ternary, and fixed-point number systems. Binary or ternary uses extremely low-bit weight representation for DNNs, which can achieve very high inference computing efficiency by eliminating multiplications, but sacrifice accuracy. Representative binary quantization methods include Binaryconnect ~\cite{courbariaux2015binaryconnect}, Binarized Neural Network (BNN) ~\cite{courbariaux2016binarized}, XNOR-net ~\cite{rastegari2016xnor}, and ABC-Net ~\cite{lin2017towards}. 
Ternary quantization schemes are implemented in TWN ~\cite{li2016ternary}, TTQ ~\cite{zhu2016trained}, and  ~\cite{he2019simultaneously}.

On the other hand, fixed-point quantization uses more bits of weight representation to preserve accuracy. For example, compared with the floating-point (e.g. 32 bits), fixed-point quantization can use 4-bit to represent the weights with negligible accuracy loss. Fixed-point quantization scheme has been implemented with different methods/algorithms. DoReFa-Net ~\cite{zhou2016dorefa} first explored it by introducing hyperbolic tangent transformation to weights and activations, with scaling factors to minimize quantization error. PACT ~\cite{choi2018pact}  improved this method by adding a parameterized clipping threshold to activations. DSQ ~\cite{gong2019differentiable} developed an evolving training method to gradually approaximate STE. QIL ~\cite{jung2019learning} parameterized the quantization interval and trained it with task loss, avoiding access to the original training data. $\mu$L2Q ~\cite{cheng2019uL2Q} introduced data distribution loss during training to minimize quantization error. LSQ ~\cite{esser2019learned} proposed a differentiable method to learn the quantizer for each layer jointly with parameters. 

Fixed-point quantization still needs multiplication operations, which execute on DSPs of FPGA. However,the DSP resources are limited, e.g., ranging from 240 to 1,540 DSP slices in Xilinx Kintex-7 series, which becomes the bottleneck of merely employing fixed point quantization.


\subsubsection{Non-Linear Number System}

Miyashita et al. ~\cite{DBLP:journals/corr/MiyashitaLM16} first replaced fixed-point quantizer with logarithmic representation to exploit bit shift operations and accelerate inference.
We refer this kind of non-linear number system as power-of-two (PoT) quantization. 
In this way, the multiplication of input (a fixed-point number) and weight (a PoT number) can be replaced by bit shift operation and can be executed as shown below, 
\begin{equation}
\begin{aligned}
2^b\times a=\begin{cases}
a<<b, \quad &b>0\\
a, \quad &b=0\\
a>>b, \quad &b<0\\
\end{cases}
\end{aligned}
.\end{equation}
where $2^b$ is the quantized weight, $a$ is the input value. 

Followed the PoT scheme, INQ ~\cite{DBLP:journals/corr/ZhouYGXC17} split weights into groups and iteratively quantize the model to low bit-width. Leng et al. ~\cite{leng2018extremely} employed ADMM training technique to increase the accuracy of extremely low bit-width DNNs. Li et al. ~\cite{li2019additive} introduced a reparameterizaiton of clipping function to get better-defined gradients, and employed weight normalization to stablize training.

Even though PoT can reduce the computation by replacing the multiplication to bit shift operation, the distance between the quantization levels grows up exponentially, causing PoT suffered from significant accuracy degradation. 
~\cite{li2019additive} proposed additive power-of-two (APoT) to reduce the accuracy degradation of power-of-Two. 
However, APoT uses more bit shift and addition while increasing the bit width for the weight quantization. For example, APoT uses 3 bit-shift and 2 addition for 6-bit weight representation, which only pursues the less accuracy degradation but does not consider the hardware availability ratio.
Besides, APoT alternatively assigns the value of Power-of-two to each part (e.g. in 5-bit quantization, APoT assigns $\{0, 2^{-1},2^{-3},2^{-5}\}$ to the first part and $\{0,1,2^{-2},2^{-4}\}$ to the second part), which is hard to deploy the quantized DNN models for hardware implementations.

\subsubsection{Bit Width Selection}
The majority of previous works ~\cite{zhou2016dorefa,choi2018pact,zhang2018lq} etc. assign same bit width to all layers, which we refer as single-precision quantization. The other track ~\cite{dong2019hawq,shen2020q} optimizes bit width for each individual layer so that maximum compression rate can be achieved with minimum degradation of accuracy. We refer this methodology as multi-precision quantization. It is to solve Hessian matrix to determine bit width for each layer. The general idea is to assign more bits to layers that are sensitive to quantization error. In fact, very few works are strictly single-precision quantization, because the first and last layers affect much more on accuracy. Therefore, current researches follow ~\cite{han2015learning} to quantize first and last layer to a higher bit width (e.g., 8 bit) or even leave them unquantized. 
Though first and last layers only compose a small percentage of computation cost, if unquantized, specific configurations are needed on FPGA to execute this different bit-width  GEMM (Generalized Matrix Multiplication), which downgrades hardware utilization.



\subsection{Motivation}

FPGA performance is constrained by the amount available DSP resources which is required for fixed-point computations. In general, inference of a fixed-point quantized DNN model is much slower than that from PoT quantization.
PoT scheme can replace the multiplications with bit-shift and addition operations that are performed by LUT resources on FPGA, but it does not fit with the true weight distribution because of its high resolution around zero and low resolution on the tails, as shown in Figure~\ref{fig:distributions_sp2}. Specifically, 4-bit PoT quantization would suffer from at least $1-2\%$ accuracy degradation.
This motivates us to adopt an improved version of PoT i.e., sum-of-power-of-two (SPoT) to mitigate the accuracy degradation. However, only relying on SPoT quantization does not exploit DSP resource on FPGA. To achieve the optimal utilization of on-chip computing resources (both DSPs and LUTs), we propose to utilize a combination of SPoT quantization and fixed-point quantization, called \emph{mixed-scheme (MS) quantization}, applying the two quantization methods respectively to different filters within a DNN layer. MS allows another dimension to choose appropriate scheme according to weight distribution, which is beneficial to accuracy.

We also observe that existing works do not quantize or use no less than 8 bits for fixed point weight representation for the first and last layers~\cite{courbariaux2015binaryconnect,he2019simultaneously,ren2019admm}, and most multi-precision works ~\cite{dong2019hawq,shen2020q} also employ such inter-layer flexibility. The deployment of these models on FPGA for inference needs different configurations for different layers. 
So we explore a novel \emph{multi-precision (MP) quantization} to quantize the first and the last layer while preserving the accuracy. Based on the optimal ratio given from hardware (FPGA) resources, we propose the our \emph{mixed-scheme, multi-precision (MSP) quantization} enjoying two benefits, i.e., (1) better utilization of the FPGA resources of FPGA, which is coming from mixed-scheme, and (2) zero accuracy degradation, which is due to multi-precision.

%% file: Section/3_SP2.tex
\section{Quantization Formulation} 
\label{sec:sp2}

In this section, we first formulate typical fixed point and power-of-two quantization schemes. Then we improve the vanilla PoT with a summation method, namely sum-of-power-of-two (SPoT) to better fit weight distribution so as to preserve accuracy with minor computation overhead introduced. Lastly we give the algorithm to perform quantization aware training. 

\subsection{Fixed Point Quantization}

As mentioned in Section \ref{sec:background}, the most intuitive way to perform quantization is to map the full precision parameters to low bit width uniform representations. The weight representation can be defined as follows: 
\begin{equation}\label{eq:fixedpointQL}
\mathcal{Q}^{FP}(m, \alpha)= \pm\alpha \times \{0, \frac{1}{2^{m-1}-1}, \frac{2}{2^{m-1}-1}, ...,  1\}.
\end{equation}
where $\mathcal{Q}^{FP}$ refers to quantized numbers, $m$ is the bit width and $\alpha$ is a scaling factor. And the mapping function from a 32-bit floating-point weight $w$ into the quantized weight $\hat w$ by $m$-bit fixed-point representation is as follows:
\begin{equation}\label{eq:fixedpointquantizer}
\begin{aligned}
\hat w  &= \prod_{\mathcal{Q}^{FP}(m, \alpha)} w\\
 &=\alpha\cdot h^{-1}\big(\frac{1}{2^{m}-1} round((2^{m}-1) \cdot h(\lceil w,\alpha\rfloor))\big),
 \end{aligned}
\end{equation}
where $\prod_{\mathcal{Q}^{FP}(m,\alpha)}(\cdot)$ denotes the quantizer function to project onto $\mathcal{Q}^{FP}(m,\alpha)$; the function $h(\cdot)$ transforms a value within $[-1,+1]$ into the range of $[0,1]$, for example we can use $h(\cdot)=\text{tanh}(\cdot)/2+0.5$; and $\lceil w,\alpha\rfloor$ clips $w$ according to
\begin{equation}
    \begin{aligned}
    \lceil w,\alpha\rfloor = 
    \begin{cases}
    -1, \quad &w<-\alpha\\
    w/\alpha, \quad &-\alpha\leq w \leq\alpha\\
    1, \quad & w>\alpha
    \end{cases}
    \end{aligned}
.\end{equation}

\subsection{Power of Two (PoT) Quantization}

In order to replace multiplications with bit shifting operations, we need to make weights in the form of powers-of-two. 
The quantized weight values by PoT scheme with an m-bit weight representation are as follows:
\begin{equation}\label{eq:P2QL}
Q^{P2}(m, \alpha)= \pm\alpha\times \{0, \frac{ 1}{2^{2^{m-1}-2}}, \frac{ 1}{2^{2^{m-1}-3}}, ...,  1\}
.\end{equation}

The mapping from continuous parameters to PoT number system is defined by
\begin{equation}\label{eq:powerof2quantizer}
\begin{aligned}
\hat w
& = \prod_{\mathcal{Q}^{P2}(m, \alpha)} w\\
& =\begin{cases}
\begin{aligned}
\alpha \cdot h^{-1}\big(2^{round(\log_2 h')}\big), \ h'>2^{-2^m+1} \\
0, \ h'\leq 2^{-2^m+1}, \\
\end{aligned}
\end{cases} \\
h' & = h(\lceil w,\alpha\rfloor), \\
\end{aligned}
\end{equation}
from which we can infer another disadvantage of pure PoT quantization. Increasing bit width merely increases resolution around mean area, but has no effect at the tail, as displayed in Figure~\ref{fig:distributions_sp2}. 
\subsection{Sum of Power of Two (SPoT) Quantization}

\begin{figure}[t]
\centering  
\includegraphics[width=0.5\columnwidth]{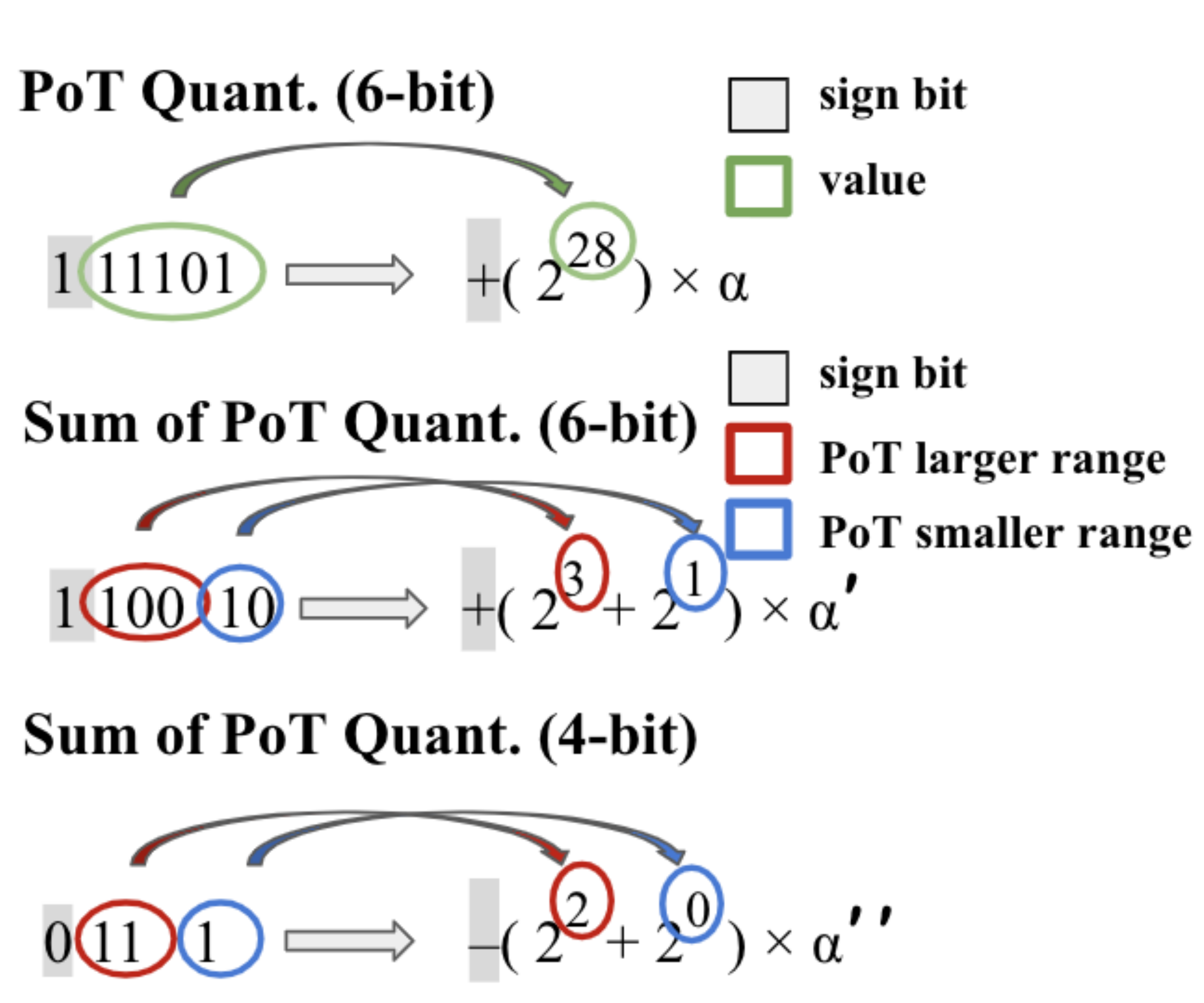}
\caption{Illustration  of  the  PoT  and  the  proposed SPoT quantization schemes. It also shows the corresponding quantization levels along with the weight distribution of an actual layer in MobileNet-v2.}
\label{fig:distributions_sp2}
\end{figure}

Sum-of-power-of-two (SPoT) can be considered as an improved version of power-of-two (PoT) quantization, which can also replace the multiplications with bit-shift operations. The quantization levels are defined as follows:
\begin{equation}\label{eq:SP2QL}
\begin{aligned}
& \mathcal{Q}^{SPoT}(m, \alpha)= \mathcal{Q}^{P2}_1(m_1, \alpha) + \mathcal{Q}^{P2}_2(m_2, \alpha)
\end{aligned}
\end{equation}
Where $m_1$ has a larger range and $m_2$ has a smaller range ($m_1\geq m_2$). Then all quantization levels set are a combination of lower bit-width PoTs. In Figure \ref{fig:distributions_sp2} we take the 6-bit SPoT quantization as an example. First, SPoT needs 1 bit for the sign, then we split the rest 5 bit to the smaller range $m_1$ (3bit) and the larger range $m_2$ (2bit). In Figure \ref{fig:distributions_sp2}, number $
0.625$ can be represented by $2^{-3} +2^{-1}$, then $2^{-3}$ can be decoded in "011" in smaller range $m_1$ and $2^{-1}$ can be decoded in "10" in larger range $m_2$. Therefore, when multiplying an input value by
weight "101101", we shift the input right 3 bit and 1 bit, respectively,
and sum the two results up.

Figure \ref{fig:distributions_sp2} also shows a notable shortcoming of PoT. PoT has very high precision around the mean, but the tail ends present very low precision. It causes a mismatch with the weight value distribution. Thus, PoT suffers from non-negligible accuracy degradation. On the other hand, the proposed SPoT has relatively even quantization intervals, which is close to that of fixed-point quantization levels. Therefore, the SPoT can achieve similar accuracy performance as the fixed-point quantization scheme.

\subsection{Quantization-Aware Training Algorithm}

Quantization is a projection from the continuous value to discrete number system, which makes the gradients flowing from loss function zero everywhere during backpropagation. There are two approaches to address this issue. One is employing a Straight Through Estimator (STE) ~\cite{DBLP:journals/corr/BengioLC13,DBLP:journals/corr/abs-1903-05662} to set the gradient to constant value $1$ as 
\begin{equation}
\begin{aligned}
&\textbf{Forward}: y=round(x) \\
&\textbf{Backward}: \frac{\partial y}{\partial x}=\textbf{1}_{x\in R}
\end{aligned}
\label{eqn:STE}
\end{equation}
The other is the Alternating Direction Method of Multipliers (ADMM)   ~\cite{leng2018extremely} to iteratively solve the parameters with a target quantization scheme as the optimization constraint. These two methods are equivalent in terms of convergence, but ADMM algorithm shows more flexibility as gradients can not be defined appropriately by STE in some specific tasks. In this paper, we perform ADMM for weight and STE for activation quantization. We will not include detailed derivations of ADMM due to space, please refer to the mentioned previous work. 

%% file: Section/4_MPQ.tex
\section{MSP Framework}\label{sec:mpq}

\subsection{Mixed-Scheme (MS) Quantization}

Relying on SPoT quantization only or fixed point quantization only will not achieve the optimal performance on FPGA devices. 
Thus, we propose the \emph{mixed-scheme- quantization (MS)}. In each layer of a DNN, we split parameters into two parts, one uses SPoT while the other is quantized under fixed point quantization scheme. 
Besides, on the algorithm level, in each layer, the weight matrix can be obtained by transforming the weight tensor into a 2D GEMM matrix. The distribution of weights in different rows is rather random. For rows that have smaller variances (have more Gaussian-like weight distributions), SPoT scheme is a better fit; while for rows with larger variances (more uniform-like distribution), using fixed-point scheme can avoid high quantization error. 
In our work, the optimal ratio of SPoT to fixed-point is determined by available resources on FPGA devices, instead of serving for accuracy. Usually, the utilization of DSPs needs to be maintained at 100$\%$ to take full advantage of the DSP resource. Incorporating with DSP, we would like to assign appropriate workload on LUTs and make them finish simultaneously, and therefore enhance the throughput.

\subsection{Multi-Precision (MP) Quantization}
\begin{figure}[t]
\centering  
\includegraphics[width=0.7\columnwidth]{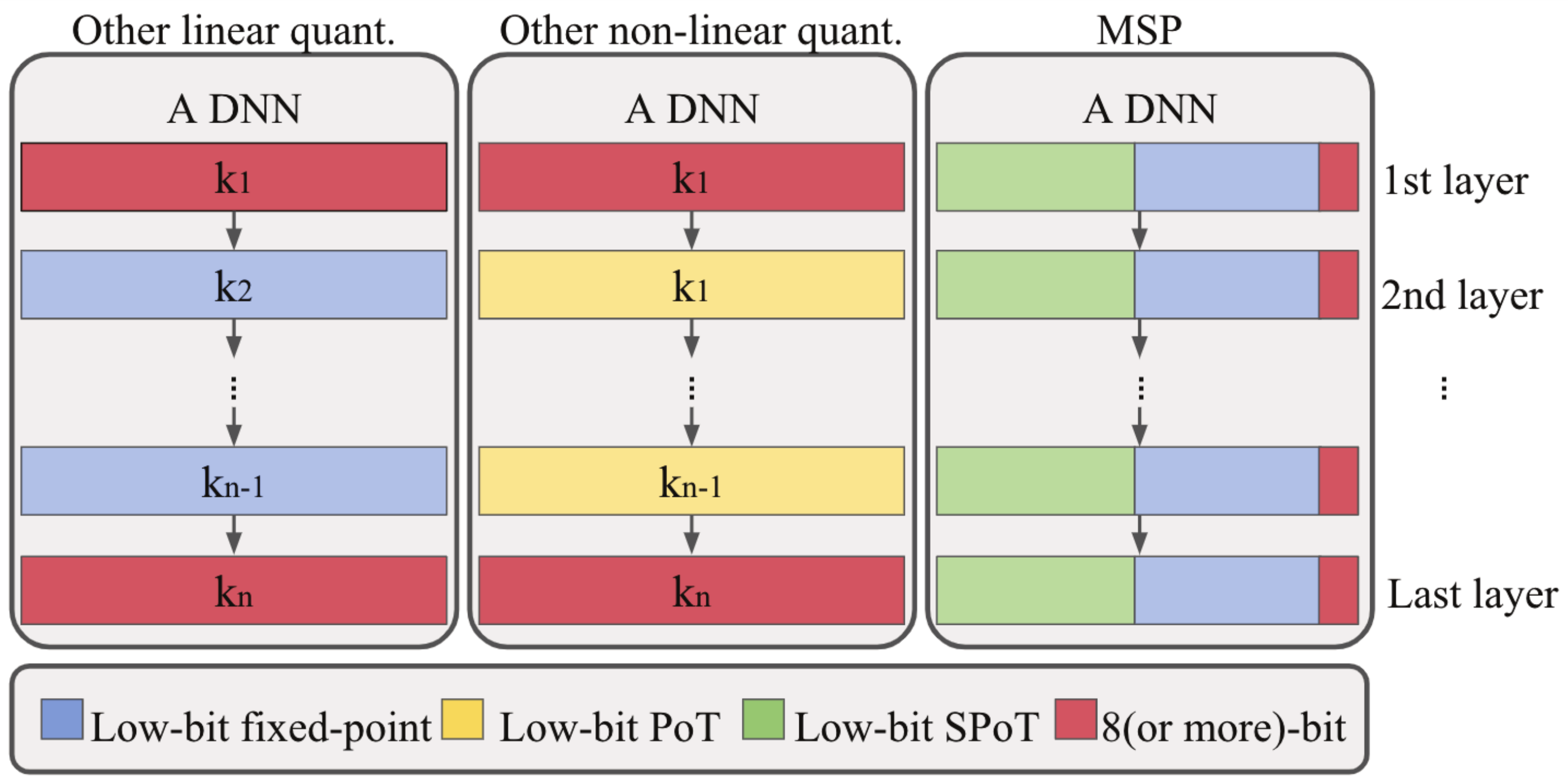}
\caption{Illustration of a DNN integrating the proposed intra-layer flexibility with the FPGA-specific quantization scheme and comparison with the other state-of-the-art quantization works. $k$ refers to the bit width of each layer. Generally, the middle layers share same bit width (e.g. $k_2=k_3=...=k_{n-1}$), but for multi-precision works, different bit widths are employed in the middle layers (e.g. $k_i\neq k_j$).} 
\label{fig:mpq}
\end{figure}

We further propose a novel mixed-precision quantization scheme (MP) to achieve lossless performance compared with the original full precision DNN models, as illustrated in Figure~\ref{fig:mpq}.
In image classification task, first and last layers are extremely sensitive.
Thus, most of the existing works do not quantize or use no less than 8-bit fixed-point weight representation for the first and last layers to maintain accuracy ~\cite{courbariaux2015binaryconnect,he2019simultaneously,ren2019admm}. 
Recent work \cite{zhu2019configurable, gong2019mixed} has investigated FPGA-based inference engine supporting different quantization bits adaptively. However, such online reconfiguration ability inside each PE incurs non-negligible hardware overhead.  Besides, such inter-layer flexibility in quantization bits brings about only minor accuracy improvement, even if sophisticated search method for per-layer quantization bits is employed ~\cite{wang2019haq, lou2019autoq}. 


To overcome this challenge, we propose \emph{Multi-precision quantization (MP)} for our quantization scheme. In each layer, we preserve the $5\%$ weights to use the 8-bit weight representation. 
Based on the average distance between the weight and the nearest 4-bit weight quantization level in each row (i.e. average quantization error) of the weight matrix. We determine the rows with highest $5\%$ average quantization error to be 8-bit, while still using the 4-bit weight representation for the rest. The overall DNN accuracy can be maintained as long as $5\%$ of weights in each layer are quantized using 8 bits. Even if the rest of the weights (in all layers) are quantized using very few bits. This is because in \emph{intra-layer flexibility}, the weights quantized using 8 bits can be trained to mitigate the imprecision caused by those weights quantized using fewer bits. This mitigation happens in every layer. On the other hand, in the prior inter-layer flexibility, the majority of layers will be quantized with fewer bits. The resultant imprecision cannot be mitigated within a layer and will be accumulated across layers. It is difficult to recover the accumulated imprecision by limited layers quantized with more bits.

Besides algorithm-level advantages, the proposed intra-layer flexibility also exhibits an advantage at the FPGA hardware level. Recall that the same quantization scheme (e.g., 4-bit for $95\%$ of weights and 8-bit for the rest of $5\%$) is applied to all layers of a DNN. At FPGA configuration time for a specific DNN inference task, one could allocate a portion of PEs for the low-bit portion of computation and the rest of PEs for the 8-bit portion, and this works for every layer. As for traditional inter-layer multi-precision scheme, it's almost impossible to perform online reconfiguration, that is, the PEs assigned to execute 8-bit first/last layers is vacant while processing the middle layers. 

\subsection{Mixed-Scheme, Multi-Precision (MSP) Quantization Framework}

To fully utilize all on-chip resources, we propose Mixed Scheme, Multi-Precision Quantization (MSP) framework as the combination of MS and MP. Based on the optimal ratio of SPoT/Fixed-point given by the hardware (FPGA) resource (e.g. 2:1 for XC7Z045). We futher give the optimal ratio of SPoT/Fixed-point/8bit is 65:30:5, which can (1) better utilize the resource of the FPGA, (2) have the highest throughput (GOPS)/lowest latency, and (3) achieve lossless accuracy performance compared with the original full precision model. The detailed implementation is shown in Algorithm \ref{algo:MSQQuantization}.

%% file: Section/5_results.tex
\section{Evaluation}\label{sec:eva}

\subsection{Experiment Setup}
\begin{table}[ht]
\small
\centering
\tabcolsep 8pt
\renewcommand\arraystretch{1.1}
\scalebox{0.8}{
\begin{tabular}{c c c c c c}
\toprule

Quantization & Bit width & \multicolumn{2}{c}{ResNet-18 Accuracy (\%)} & \multicolumn{2}{c}{MobileNet-v2 Accuracy (\%)} \\
Scheme & (Wght./Actv.) & Top1 (N/Y) & Top5 (N/Y)& Top1 (N/Y) & Top5 (N/Y)\\
\hline
\multicolumn{6}{c}{\textbf{CIFAR-10}} \\
\hline
Baseline  & 32/32 & 93.62 & - & 92.51 & -\\
PoT & 4/4 &92.97
/ 92.14& - & 91.34 / 90.92&-\\
Fixed & 4/4 & 93.43 / 92.97 & - & 92.34 / 91.76 & -\\
SPoT & 4/4 & 93.47
/ 92.94 & - & 92.72
/ 91.83& -\\
MS& 4/4 & 93.53
/ 92.98& - & 92.57
/ 91.99& -\\
MP & 4/4 & - / 93.63
&-  & - / 92.54
&- \\
\bf{MSP} & 4/4 & - / \bf{93.72}
&-  & - / \bf{92.58}
&- \\
\hline
\multicolumn{6}{c}{\textbf{CIFAR-100}} \\
\hline
Baseline & 32/32 &  74.49&  92.70& 71.48& 91.98\\
PoT & 4/4 &73.88
/ 72.97 &92.14
/ 91.65 &68.68
/ 67.11& 90.06
/ 89.21\\
Fixed & 4/4 &  74.37
/ 73.88&  92.31 / 91.72& 71.16
/ 70.22 & 91.63
/ 90.88\\
SPoT & 4/4 &  74.33
/ 73.97&  92.49
/ 92.03 & 71.13
/ 70.21& 91.69
/ 90.85\\
MS & 4/4 & 74.58
/ 74.03& 92.51
/ 92.05 & 71.21
/ 70.25& 91.74
/ 90.92 \\
MP & 4/4 & - / 74.54
& - / 92.61
& - / 71.49
& - / 91.82 \\
\bf{MSP} & 4/4 & - / \bf{74.61}
& - / \bf{92.69}
& - / \bf{71.51}
& - / \bf{91.97} \\
\hline
 & &\multicolumn{2}{c}{\textbf{ImageNet}} & \multicolumn{2}{c}{Wght./Actv. 4/32} \\
\hline
Baseline & 32/32 & 69.76 & 89.08 & 71.88& 90.29\\
PoT &4/4 &68.20
/ 67.11 &87.14
/ 85.93& 69.93 / 67.88& 88.63 / 86.83\\
Fixed & 4/4 & 69.72
/ 68.66 &  88.67
/ 87.54& 71.26
/ 69.23& 90.18 / 88.03 \\
SPoT& 4/4 & 69.74
/ 68.48 &  88.71
/ 87.92 & 71.32
/ 69.76& 90.17
/88.42\\
MS & 4/4 &  70.11
/ 69.22 & 89.41
/ 88.33& 71.26
/ 69.31& 90.04
/ 88.11\\
MP & 4/4 & - / 69.99 
& - / 89.12 
& - / 71.68 & - / 90.22\\
\bf{MSP} & 4/4 & - / \bf{70.47} 
& - / \bf{89.52} 
& - / \bf{71.73} & - / \bf{90.27}\\

\bottomrule
\end{tabular}
}
\caption{Result from different quantization schemes for the ResNet-18 and MobileNet-v2 DNN models on CIFAR-10, CIFAR-100, and ImageNet datasets. Y/N: With/Without quantization of the first and the last layer.}
\label{tab:comparison_schemes}
\end{table}

\begin{table}[t]
\small
\centering
\tabcolsep 2.2pt
\begin{tabular}{ccccc}
\toprule
\multirow{2}{*}{Methods} & Top-1 & Top-5 &
quant. &
quant.\\
~  & (\%) & (\%) &1st $l$.& Last $l$.\\
\hline
Baseline  & 69.76 & 89.08& -& -\\
Dorefa ~\cite{zhou2016dorefa}  & 68.10 & 88.10&$\times$&$\times$\\
PACT ~\cite{choi2018pact} & 69.20 & 89.00&$\times$&$\times$\\
DSQ ~\cite{gong2019differentiable}  & 69.56 & N/A&$\times$&$\times$\\
QIL ~\cite{jung2019learning}  & 70.10 & N/A&$\times$&$\times$\\
$\mu$L2Q ~\cite{cheng2019uL2Q}  & 65.92 & 86.72&-&16bit\\
LQ-NETS ~\cite{zhang2018lq} & 69.30 & 88.80&$\times$&$\times$\\
\textbf{MSP (ours)}  & \textbf{70.47} & \textbf{89.52}&\textbf{\checkmark}&\textbf{\checkmark}\\

\bottomrule
\end{tabular}
\caption{Comparisons with existing works with ResNet-18 model on ImageNet dataset for 4 bit quantization.} \label{tab:imagenetresnet}
\end{table}

\begin{table}[t]
\small
\centering
\tabcolsep 3pt
\renewcommand\arraystretch{1.1}
\scalebox{0.7}{
\begin{tabular}{ccccccccccccccccc}
\toprule
 &  & \textbf{MSP} & MSP & MSP & MP & MS & MS & MS &Fixed & Fixed & PoT & PoT & SPoT & SPoT \\
 \hline
\multirow{2}{*}{DSP} & 8-bit FP & 5\% & 5\% & 10\% & 5\% & 0 & 0 & 0 & 0 & 0 & 0 & 0 & 0 & 0 \\
 & 4-bit FP & 30\% & 0 & 0 & 95\% & 50\% & 50\% & 33\% & 100\% & 100\% & 0 & 0 & 0 & 0 \\
\multirow{2}{*}{LUT} & 4-bit SPoT & 65\% & 95\% & 90\% & 0 & 50\% & 50\% & 67\% & 0 & 0 & 0 & 0 & 100\% & 100\% \\
 & 4-bit PoT & 0 & 0 & 0 & 0 & 0 & 0 & 0 & 0 & 0 & 100\% & 100\% & 0 & 0 \\
\multicolumn{2}{c}{First and Last Layer (Y/N)} & Y & Y & Y & Y & N & Y & N & N & Y & N & Y & N & Y \\
\hline
\multicolumn{2}{c}{Top-1 Accuracy (\%)} & \textbf{70.47} & 70.09 & 70.24 & 69.99 & 70.08 & 69.52 & 70.11 & 69.72 & 68.66 & 68.20 & 67.11 & 69.74 & 68.48 \\
\multicolumn{2}{c}{FPGA Latency (ms)} & \bf{11.2} & 13.7 & 13.3 & 29.1 & 20.1 & 12.2 & 15.4 & 39.5 & 25.4 & 19.5 & 14.7 & 22.9 & 16.5 \\
\bottomrule
\end{tabular}
}
\caption{Ablation study of ResNet-18 on ImageNet. Y/N: With/Without quantization of the first and the last layer. The results are measured on XC7Z045.}
\label{tab:ablation}
\end{table}

We evaluate our novel MSP on image classification tasks with convolutional neural networks (CNNs). We use no extra data augmentations other than those already employed for training the 32-bit floating-point baseline models. Our quantization training algorithm uses step or cosine learning rate decay and $\ell_2$ regularization, following training algorithms of the baseline models. Our experiments are implemented with server-grade machines running Ubuntu 18.04, CUDA
10.2 and PyTorch 1.5. Our models are trained on NVIDIA TITAN RTX GPUs and GeForce RTX 2080Ti GPUs.
We evaluate with the deep residual net (ResNet-18)~\cite{he2016deep}, which generalizes well for varieties of tasks, as well as the lightweight MobileNet-v2 model ~\cite{sandler2018mobilenetv2}.
We test on CIFAR-10, CIFAR-100 ~\cite{krizhevsky2009learning}, and ImageNet ILSVRC-2012 ~\cite{krizhevsky2012imagenet} datasets.
DNN models for CIFAR-10 and CIFAR-100 datasets are trained from scratch, and quantized for $150$ epochs.
For ImageNet dataset, pre-trained models in 32-bit floating-point are used, and quantized for $90$ epochs. 
The initial learning rates are $8e-3$ for CIFAR-10, $4e-3$ for CIFAR-100, $1e-2$ for ImageNet.

To demonstrate hardware efficiency of the proposed MSP framework, we implemented the architecture with heterogeneous $GEMM$ cores on the embedded FPGA device, in which high efficiency is usually top priority under resource constraints. We validate our framework on Zynq XC7Z020 and XC7Z045 devices with different design parameters that result in different throughput and resource utilization results.
We setup different ratios between the PE array sizes of the $GEMM_{SPoT}$,$GEMM_{fixed}$, and $GEMM_{8-bit}$ cores. 
Specifically, we progressively increase the size of $GEMM_{SPoT}$ core ($Blk_{out,SPoT}$), till the LUT utilization reaches the highest ratio.
For all implementations, the working frequency is set to 100MHz.

\subsection{Accuracy Performance}

Tables \ref{tab:comparison_schemes} and \ref{tab:imagenetresnet} summarize quantization results for image classification task.
We use $PR_{SPoT:Fixed:8-bit}=65:30:5$, which is the optimal ratio validated from resource utilization results on FPGA. MSP obtains the minimum accuracy degradation.

The accuracy increase of MSP compared to PoT, Fixed, SPoT, MS or MP results from several aspects. First, combining SPoT and Fixed makes the quantized DNN weights fit the original weight distribution better. Besides, leave $5\%$ weight with 8-$bit$, which is "intra-layer flexibility" can achieve lossless performance even the first and the last layer are quantized.
In addition, model compression could slightly increase accuracy when weight bit-width $\geq 4$, as 
quantization noise can potentially act as regularization that benefits generalization and addresses overfitting.

Tables \ref{tab:imagenetresnet}  compares our MSP with existing DNN quantization works including Dorefa ~\cite{zhou2016dorefa},
PACT ~\cite{choi2018pact},
DSQ ~\cite{gong2019differentiable},
QIL ~\cite{jung2019learning},
$\mu$L2Q ~\cite{cheng2019uL2Q}, and
LQ-NETS ~\cite{zhang2018lq}.
Those works and our MSP start with the same pre-trained models with the same baseline accuracy.
Table \ref{tab:imagenetresnet} shows that Dorefa, PACT, DSQ, $\mu$L$2$Q, and LQ-NETS have up to $3.84\%$ accuracy degradation, only QIL reports lossless accuracy performance. Our MSP increases accuracy by $0.71\%$ compared with the floating-point model. Note that those works above they do not quantize the first and the last layer, only $\mu$L2Q uses 16bit weight representation for the last layer. On the other hand, MSP quantizes the first and the last layer while still can achieve the lossless accuracy performance.

\subsection{Hardware Efficiency}

To present the performance with real-world applications, we employed different CNN models with the proper SPoT/fixed-point/8-bit ratios on the two devices. The networks ResNet-18 and MobileNet-v2 are implemented based on the ImageNet dataset. The performance results of each network under various hardware configurations are displayed in Table \ref{tab:perf}. 
Generally, the utilization of DSP is always $100\%$, by increasing the ratio of SPoT, the utilization of the other resource has improved (e.g. from $19\%$ to $69\%$ on LUT in XC7Z045).
On the other hand, the heterogeneous $GEMM_{SPoT}$, $GEMM_{fixed}$, and $GEMM_{8-bit}$ cores improve the throughput by $2.2\times-2.7\times$ with the optimal design compared to utilizing the $GEMM_{fixed}$ core only
(e.g. the throughput from 120.5 to 325.0 GOPS on ResNet-18 in XC7Z045). 

\subsection{Ablation Study}
We experiment over different quantization configurations to validate the proposed MSP scheme. Note that all the percentages refer to the ratio of employed quantization scheme in the main body of a DNN, since first and last layers are specified separately. As shown in table \ref{tab:ablation}, single scheme quantization (fixed point ~\cite{zhou2016dorefa}, PoT, SPoT) shows no advantage in both accuracy and speed (FPGA latency in our work). Plus, simply quantizing first and last layer into low bit width introduces noticeable accuracy degradation. With our Mixed Scheme (MS) method, execution speed is increased significantly because of efficient utilization of both DSP and LUTs. We can also observe slight improvement in accuracy, because the dual schemes are organized to better fit weight distribution, as described in \ref{sec:sp2}. Finally we employ our intra-layer mixed precision (MP) to mitigate the need of high bit widths in first and last layers, further improving speed and achieving comparable accuracy. 

%% file: Section/6_conclusion.tex
\section{Conclusion}\label{sec:con}

This work proposes a mixed-scheme (MS) quantization method that combines the sum-of-power-of-2 (SPoT) quantization scheme 
and the traditional fixed-point quantizer.
Multiplications under the SPoT scheme can be replaced with bit shifting and addition operations using the FPGA LUT resources,
and the fixed-point operations can be executed efficiently on DSP modules, therefore the heterogeneous resources on FPGA are fully exploited. 
Furthermore, we also develop a novel intra-layer multi-precision (MP) method to mitigate the need of high bit width in first and last layers while still preserving accuracy. As a new dimension for mixing bit width, our MP is much more hardware friendly compared to inter-layer multi-precision quantization works. 
Finally, we combine the MS and MP to MSP, as an ensemble of SPoT ,low-bit fixed point and $5\%$ 8bit quantization. MSP achieves best accuracy results as well as fastest inference speeds on FPGA compared to prior arts.
With optimal SPoT/fixed-point/8bit ratios, our FPGA-specific quantization scheme can not only achieve $70.47\%$ Top1 accuracy in ResNet-18 on the ImageNet dataset but also speedup $3.53\times$ inference time compared with pure fixed point quantization on FPGA devices. 